# FactLLaMA: Optimizing Instruction-Following Language Models with External Knowledge for Automated Fact-Checking

Tsun-Hin Cheung and Kin-Man Lam

Department of Electrical and Electronic Engineering, The Hong Kong Polytechnic University, Kowloon, Hong Kong

E-mail: tsun-hin.cheung@connect.polyu.hk, enkmlam@polyu.edu.hk

*Abstract* — **Automatic fact-checking plays a crucial role in combating the spread of misinformation. Large Language Models (LLMs) and Instruction-Following variants, such as InstructGPT and Alpaca, have shown remarkable performance in various natural language processing tasks. However, their knowledge may not always be up-to-date or sufficient, potentially leading to inaccuracies in fact-checking. To address this limitation, we propose combining the power of instruction-following language models with external evidence retrieval to enhance fact-checking performance. Our approach involves leveraging search engines to retrieve relevant evidence for a given input claim. This external evidence serves as valuable supplementary information to augment the knowledge of the pretrained language model. Then, we instruct-tune an open-sourced language model, called LLaMA, using this evidence, enabling it to predict the veracity of the input claim more accurately. To evaluate our method, we conducted experiments on two widely used fact-checking datasets: RAWFC and LIAR. The results demonstrate that our approach achieves state-of-the-art performance in fact-checking tasks. By integrating external evidence, we bridge the gap between the model's knowledge and the most up-to-date and sufficient context available, leading to improved fact-checking outcomes. Our findings have implications for combating misinformation and promoting the dissemination of accurate information on online platforms. Our released materials are accessible at: https://thcheung.github.io/factllama.**

## I. INTRODUCTION

With the continuous growth of social media and online communication, we are experiencing a deluge of information through these channels, raising the urgent need to distinguish facts from fiction. Unfortunately, these platforms provide almost everyone with the ability to spread misinformation or fake news that often goes viral, leading to the widespread dissemination of inaccurate or false information to a global audience. Misinformation can take different forms, ranging from fabricated news stories to manipulated images or videos aimed at swaying public opinion [1].

To counteract this alarming trend, the research community is continuously developing new and innovative approaches to tackle the problem of identifying and correcting misinformation. Artificial intelligence (AI)-assisted fact-checking is one such approach and has been gaining attention in recent years [2]. It has the potential to automate the laborious and time-consuming process of manually verifying facts, while enabling quick dissemination of accurate information. With the recent rise in popularity of Large Language Models (LLMs) [3], [4] in Natural Language Processing (NLP) tasks, such as machine translation, text classification, and data extraction, they have also been used for fake news detection [5].

Despite the impressive capabilities of these LLMs, a significant limitation of these language models is their reliance on pre-existing knowledge, which may not always be up-to-date or sufficient. In the context of fact-checking, the reliance of language models solely on their internal knowledge raises concerns about their ability to accurately assess the veracity of claims, especially when faced with rapidly evolving information [6]. To address this limitation, it becomes imperative to consider external knowledge sources that provide updated and reliable information in recent fact-checking algorithms [6]–[9].

This paper aims to enhance the fact-checking capabilities of instruction-following language models by leveraging external evidence. We propose a method that combines the power of pretrained language models with the retrieval of relevant external evidence from search engines. By integrating this external evidence during the instruct-tuning process, we aim to augment the knowledge of the language model, enabling it to make more accurate predictions. The contributions of our work are summarized as follows:

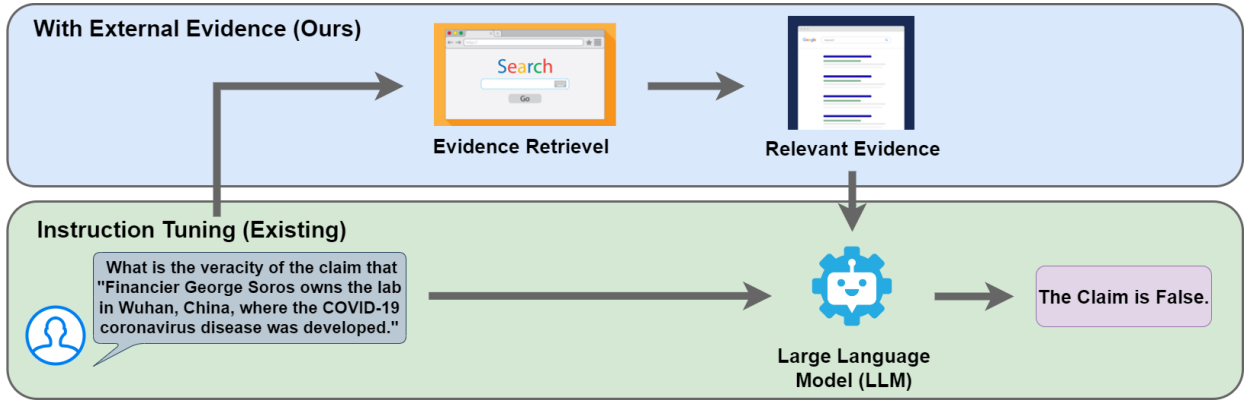

Fig. 1. Our approach for automatic fact-checking with external evidence retrieved from search engine.

- **Introducing instruct-tuned language models for fact-checking**: We propose the application of instruct-tuned language models, i.e., LLaMA, for automatic fact-checking tasks, expanding their scope beyond language generation.
- **Addressing the limitations of instruct-tuned models**: We identify the limitations of instruct-tuned language models in fact-checking due to outdated knowledge, and propose the integration of external evidence to enhance their accuracy and reliability.
- **Proposing a method for incorporating external knowledge**: We present a novel approach that combines pretrained language models with external evidence retrieval from search engines, augmenting the knowledge base for fact-checking.
- **Achieving state-of-the-art performance**: Through experiments on the RAWFC and LIAR datasets, we demonstrate that our method achieves state-of-the-art performance in fact-checking tasks.

## II. RELATED WORK

### A. Automatic Fact-Checking

In recent years, the task of automatic fact-checking has gained significant attention due to the growing concern about the spread of false information. Many studies have been carried out using different machine learning paradigms. This section provides an overview of related works that focus mainly on machine learning-based, deep learning-based, and transformer-based approaches to automated fact-checking.

Machine learning algorithms have been widely used to build models for automated fact-checking. In traditional machine learning methods, the features are manually engineered and used to train a classifier. Several studies have explored and compared different machine learning-based methods for automated fact-checking. Hassan et al. [10] proposed a solution based on the Support Vector Machine algorithm that performs a multi-level fact-checking process on news claims. Their results indicate that the proposed solution outperforms several traditional methods in terms of accuracy.

Deep learning-based methods have been successful in many Natural Language Processing (NLP) tasks, including automated fact-checking [11]. Deep neural networks can automatically learn complex features from raw data, including word embeddings and network architectures that operate on these embeddings. In 2018, Popat et al. [12] proposed a neural network-based solution to verify claims by directly comparing them to the corresponding evidence. The model utilizes a hierarchical attention mechanism to extract relevant evidence and compares it to the given claim to make a prediction. Another study presented a fact-checking system that combines deep neural networks with natural language generation to generate explanations for the model's output [9].

The Transformer is a recent innovation in deep learning that has shown great potential in several natural language processing tasks, including automated fact-checking. Transformer-based models [13], [14] employs self-attention mechanisms that enable them to capture long-range dependencies between input tokens. Recently, several studies have demonstrated the effectiveness of Transformer-based models in automated fact-checking. For instance, Kotonya et al. [6] proposed a Transformer-based solution that operates on pairs of claims and evidence. Their model utilizes an attention mechanism that allows it to focus on the most relevant pieces of evidence when verifying a claim. The authors reported that their model achieved high accuracy on a publicly available fact-checking dataset. Another study utilized a Transformer model to predict the veracity of a claim by performing joint

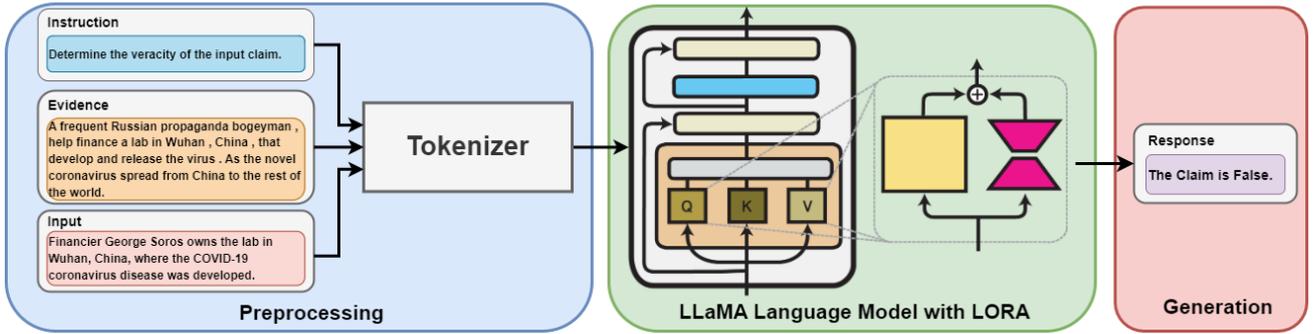

Fig. 2. Illustration of optimizing instruction-following models with external evidence using Low Rank Adaption Tuning (LORA)

reasoning over both textual evidence and commonsense knowledge [15].

### B. Instruction-Following Language Models

Instruction-following language models have emerged as a promising and novel approach for many NLP tasks [16]. These models can follow a set of instructions to perform a specific task, and perform classification and generation tasks.

InstructGPT [17] is an instruction-following language model that combines the power of the GPT architecture with a sequence of natural language instructions to perform a specific task. In the context of automated fact-checking, the model is presented with a claim. It then uses a set of instructions to perform verification for its output. The InstructGPT model has been evaluated on several publicly available commonsense reasoning datasets and has been shown to achieve high accuracy.

Self-Instruct [18] is an extension of the InstructGPT model that leverages self-supervised learning to improve its performance on a variety of natural language processing tasks. The model uses a simple heuristic algorithm that generates its own sequence of instructions and samples from the task pool. This self-supervised approach has been shown to improve the model's accuracy on many NLP tasks.

ChatGPT [19] is another instruction-following language model that can perform a wide range of natural language processing tasks, including automated fact-checking. The model leverages a conversational approach to fact-checking by generating questions that a human fact-checker would ask and answering them on its own. To achieve this, the model follows a set of instructions when interacting with input claims. New Bing from Microsoft[1] is a recently released industrial product that is powered by GPT and Bing search engine. However, both ChatGPT and New Bing are proprietary models, which cannot be tuned by the general public and researchers for downstream uses. Moreover, limited access to ChatGPT restricts the number of times that an account can perform fact-checking, especially for large-scale use.

Stanford Alpaca [20] is an open-sourced instruction-following language model based on the paradigm of open sequence-to-sequence model, called LLaMA [21], that is capable of performing a wide range of natural language processing tasks. In the context of fact-checking, the model can be used presented with a claim. The model derives a sequence of rules from the task specification and applies them to predict whether a given claim is factual or not. In our work, we aim to improve the fact-checking capability of LLaMA with the help of external knowledge.

### C. Low-Rank Adaptation Tuning

Low-Rank Adaptation (LORA) [22] is a parameter-efficient fine-tuning method that focuses on identifying the most important neurons in a pretrained language model and updating only those neurons to improve performance on a target dataset. The "Low-Rank" component in LORA involves low-rank matrix factorization in the fine-tuning process, which compresses pretrained model parameters into a lower-dimensional space, while preserving as much of the original information as possible. The resulting lower-dimensional parameter space can be fine-tuned more efficiently and effectively using fewer target samples. In our work, we leverage the LORA tuning in instruct-tuning LLaMA, because LORA tuning can reduce GPU memory to 16 GB during tuning, when compared to 64GB during fine-tuning [20].

### III. METHODOLOGY

In this section, we describe the methodology for instruct-tuning language models with external evidence for automatic fact-checking. Our proposed method consists of two key components: (1) the generation of instruction-evidence-input claim samples and (2) instruct-tuning of a generative pretrained language model, i.e., LLaMA, using these samples.

---
[1] https://www.bing.com/new

### A. Instruction-Evidence-Input Generation

**Generation of instruction-evidence-input samples**. In order to train the instruct-tuned language model, we first generate instruction-evidence-input samples. We combine the instruction, evidence, and input claim into a single sequence, with appropriate special tokens to differentiate them. The instruction provides guidance on how to incorporate the evidence for fact-checking, while the evidence consists of relevant information, retrieved from search engines, using the Google API. The input claim represents the claim that needs to be fact-checked. Similar to previous studies [8], [12], we filtered out those evidence that are from fact-checking sites or published before the claim is recorded in the fact-checking sites.

**Evidence collection pipeline**. Our evidence collection pipeline utilizes the Google API to retrieve relevant evidence given the textual inputs. We formulate queries, based on the input claim, and use the API to search for information from reputable sources. The retrieved evidence includes sentences or paragraphs that are likely to contain relevant information related to the claim. This external evidence serves as valuable supplementary information for the fact-checking process.

Converting factual classification to sequence-to-sequence problem. To convert the factual classification task into a sequence-to-sequence problem suitable for generative transformer models, we frame the task as text generation for automatic fact-checking. We treat the input claim as the source sequence and the fact-checking result as the target sequence. The instruct-tuned language model learns to generate a fact-checking response given the input claim and the relevant evidence. This approach allows the model to capture the nuances and context of the claim and produce more accurate fact-checking outcomes.

### B. Tuning Pretrained Instruction-Following Models with External Knowledge

In this study, we instruct-tune the pretrained LLaMA [21] model using the LORA algorithm [22]. Our approach not only takes the text claim as input for factual classification, but also the retrieval evidence.

**Optimization of language model**. During instruct-tuning, we aim to optimize our LLaMA model's parameters $\theta$ to minimize a loss function that measures the difference between the predicted fact-check results and the ground truth of the training dataset. Suppose we have a set of input-output pairs $(x, y)$ in the training set, where $x$ is the instruction-evidence-input to be input to the LLaMA model, and $y$ is the corresponding label for that claim. Suppose $f(x; \theta)$ denotes the output, i.e., predicted veracity of the input claim, of the LLaMA with parameters $\theta$ for input $x$. We define the loss function as follows:

$$L(\theta) = \sum(y - f(x; \theta))^2. \quad (1)$$

**LORA tuning**. Our goal is to find the optimal values for the parameters of the LLaMA model $\theta$ that minimize the loss function $L(\theta)$. To achieve this, we leverage the LORA algorithm, which involves a low-rank approximation of the parameter matrix $\theta$. As a result, it reduces the number of trainable parameters and improves generalization, while preserving most of the information contained in the original parameters. Specifically, LORA compresses $\theta$ into a low-rank matrix product:

$$\Omega = UV, \quad (2)$$

where $U$ and $V$ are matrices of lower rank than $\theta$. The LORA algorithm updates the parameters $\theta$ by adding a regularized factorization of $\theta$ to the existing parameters, as follows:

$$\theta' = UV, \quad (3)$$

$$\theta' = \theta + (\Omega - \theta), \quad (4)$$

where $\theta'$ is the updated parameter matrix.

### IV. EXPERIMENTAL SETUPS AND RESULTS

### A. Datasets

We use two publicly available datasets, namely the RAWFC [8] and LIAR [23] datasets, for evaluating our proposed approach. The RAWFC dataset is a collection of claims related to factual verification tasks, consisting of 2,012 claims with supporting evidence, labelled either true, false, and half-true, from Snopes.com. The LIAR dataset is a dataset of political statements fact-checked by PolitiFact.com, consisting of 12,836 short statements, each labelled as true, mostly true, half-true, barely true, false, or pants-on-fire. For fair comparison to other methods, we use the same split released in [8].

### B. Experimental Setup

We run the experiments using the smallest LLaMA model, i.e., LLaMA-7B [21] model, in our experiment, as it is the largest model that fits our hardware setup. We evaluate our proposed approach for automatic fact-checking using three standard evaluation metrics: Precision Scores, Recall Scores, and F1 Scores. To instruct-tune the LLaMA model, we trained the models for 3 epochs with a mini-batch size of 32. We employ the Adam optimizer with an initial learning rate of 10-4 and a linear learning decay from the initial value to 0. To avoid overfitting, we applied a dropout rate of 0.05. Our models were implemented in PyTorch [24] and HuggingFace [25]. All experiments are conducted on two GeForce RTX 2080 Ti GPUs.

*C. Comparison to Other Methods*

Table I and Table II provide the evaluation results of various methods on two different fact-checking datasets, i.e., RAWFC and LIAR. The methods are compared based on precision, recall, and F1-score, which are commonly used metrics to assess the performance of classification tasks. The baselines that we compared include SVM [26], which utilizes bag-of-words features for fake news detection. CNN [23] incorporates metadata features to enhance representation learning. RNN [11] learns representations from word sequences without relying on external resources. DeClarE [12] combines word embeddings from the claim, report, and source to assess the credibility of the claim. dEFEND [7] employs a GRU-based model for veracity prediction, providing explanations. SentHAN [6] represents each sentence based on coherence and semantic conflicts with the claim. SBERT-FC [6] utilizes SentenceBERT (SBERT) for encoding and identifies fake news based on the top-ranked sentences. GenFE [9] and GenFE-MT [9] detect fake news independently or jointly with explanations in a multi-task setup. CofCED [8] is a Coarse-to-fine Cascaded Evidence-Distillation neural network for explainable fake news detection based on such raw reports, alleviating the dependency on fact-checked ones.

From Table I for the RAWFC dataset, it can be observed that traditional machine learning methods, like SVM, CNN, and RNN, achieve moderate results in terms of precision, recall, and F1-score. However, more advanced models, such as DeClarE, dEFEND, sentHAN, SBERT-FC, GenFE, GenFE-MT, and CofCED, outperform the traditional methods, particularly CofCED, which achieves the highest F1-score of 0.5107.

Interestingly, LLaMA without tuning, i.e., zero-shot prediction, performs relatively poorly compared to the other methods. However, when Instruct-tuning is applied, there is a significant improvement in performance, particularly when external knowledge is incorporated. Instruct-tuned LLaMA with external knowledge achieves the highest F1-score of 0.5565, surpassing all other methods and demonstrating the effectiveness of leveraging external evidence.

On the evaluation on the LIAR dataset, as shown in Table II, similar patterns can be observed. Traditional machine learning methods show relatively low performance, while more advanced models exhibit better results. CofCED achieves the highest F1-score of 0.2893, indicating its effectiveness in fact-checking on the LIAR dataset.

| Methods | Precision | Recall | F1 |
| --- | --- | --- | --- |
| SVM [26] | 0.3233 | 0.3251 | 0.3171 |
| CNN [23] | 0.3880 | 0.3850 | 0.3859 |
| RNN [11] | 0.4135 | 0.4209 | 0.4039 |
| DeClarE [12] | 0.4339 | 0.4352 | 0.4218 |
| dEFEND [7] | 0.4493 | 0.4326 | 0.4407 |
| sentHAN [27] | 0.4566 | 0.4554 | 0.4425 |
| SBERT-FC [6] | 0.5106 | 0.4592 | 0.4551 |
| GenFE [9] | 0.4429 | 0.4474 | 0.4443 |
| GenFE-MT [9] | 0.4564 | 0.4527 | 0.4508 |
| CofCED [8] | 0.5299 | 0.5099 | 0.5107 |
| LLaMA (w/o tuning) [20] | 0.3350 | 0.3255 | 0.2643 |
| FactLLaMA (Instruct-tuning w/o external knowledge) | 0.5376 | 05400 | 0.5376 |
| **FactLLaMA (Instruct-tuning with external knowledge)** | **0.5611** | **0.5550** | **0.5565** |

Table I. Results on the RAWFC dataset.

| Methods | Precision | Recall | F1 |
| --- | --- | --- | --- |
| SVM [26] | 0.1578 | 0.1592 | 0.1534 |
| CNN [23] | 0.2258 | 0.2239 | 0.2136 |
| RNN [11] | 0.2436 | 0.2120 | 0.2079 |
| DeClarE [12] | 0.2286 | 0.2055 | 0.1843 |
| dEFEND [7] | 0.2309 | 0.1856 | 0.1751 |
| sentHAN [27] | 0.2264 | 0.1996 | 0.1846 |
| SBERT-FC [6] | 0.2409 | 0.2207 | 0.2219 |
| GenFE [9] | 0.2801 | 0.2616 | 0.2649 |
| GenFE-MT [9] | 0.1855 | 0.1990 | 0.1515 |
| CofCED [8] | 0.2948 | 0.2955 | 0.2893 |
| LLaMA (w/o tuning) [20] | 0.1587 | 0.2069 | 0.1224 |
| FactLLaMA (Instruct-tuning w/o external knowledge) | 0.3232 | 0.3157 | 0.2998 |
| **FactLLaMA (Instruct-tuning with external knowledge)** | **0.3246** | **0.3205** | **0.3044** |

Table II. Results on the LIAR dataset.

Once again, LLaMA without tuning performs poorly, but instruct-tuning leads to substantial improvements.

Incorporating external knowledge in the instruct-tuning process further enhances the performance, with LLaMA, Instruct-tuning and external knowledge achieving the highest F1-score of 0.3044.

In summary, the evaluation results from both datasets highlight the superiority of advanced models over traditional machine learning methods in fact-checking tasks. The instruct-tuning approach, particularly when combined with external knowledge, consistently outperforms other methods, showcasing the value of leveraging external evidence for accurate fact-checking. These findings emphasize the importance of staying updated with the latest information and leveraging advanced techniques to effectively combat the spread of misinformation.

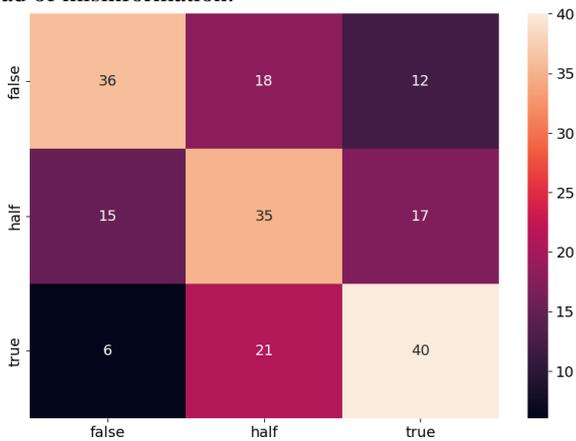

**Fig.3** Confusion matrix on the RAWFC Dataset.

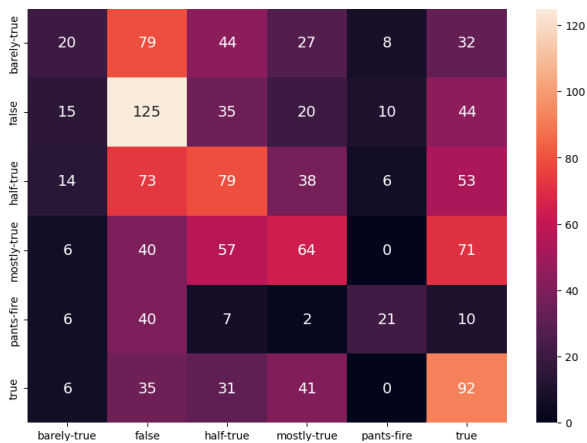

**Fig.4** Confusion matrix on the LIAR Dataset.

### D. Confusion Matrices

Figures 3 and 4 present the confusion matrices for the RAWFC and LIAR datasets, respectively. The rows and columns in the figures represent the ground-truth and predictions, respectively.

In Figure 3, it is evident that the model can effectively distinguish between the TRUE and FALSE labels. However, classifying the HALF-TRUE label proves to be more challenging for the model. This difficulty arises because both the HALF-TRUE and FALSE labels contain misinformation, albeit with differing degrees of accuracy. Moving to Figure 4, we observe that the model shows clear classification performance for the TRUE and PLANT-FIRES classes, compared to the other classes. However, it struggles to accurately classify the BARELY-TRUE, HALF-TRUE, and MOSTLY-TRUE classes. This difficulty arises from the fact that items in these classes contain a mixture of true and false information, making it a subjective task for both humans and machines to classify them accurately without specialized expertise.

## V. CONCLUSION

In conclusion, this research highlights the crucial role of automatic fact-checking in combating the spread of misinformation online. While Large Language Models (LLMs) and Instruction-Following variants, like InstructGPT and Alpaca, have demonstrated remarkable performance in various natural language processing tasks, their potential lack of up-to-date or sufficient knowledge can lead to inaccuracies in fact-checking. To address this limitation, we proposed a method that combines pretrained language models with external evidence retrieval, resulting in enhanced fact-checking accuracy. By leveraging search engines to retrieve relevant evidence for a given claim, we successfully augmented the knowledge of the pretrained language model. Through instruct-tuning an open-source language model called LLaMA, with this external evidence, we achieved more accurate predictions regarding the veracity of input claims. Experimental evaluations on widely used fact-checking datasets, RAWFC and LIAR, showcased that our approach achieved state-of-the-art performance in fact-checking tasks. The integration of external evidence effectively bridged the knowledge gap between the model and the most up-to-date information available, leading to improved fact-checking outcomes. We believe our research has significant implications for combatting misinformation and promoting the dissemination of accurate information on online platforms. In our future work, we plan to generate explanations with these pretrained language models for more general use.